\pdfoutput=1

\documentclass[11pt]{article}

\usepackage[]{acl}

\usepackage{times}
\usepackage{latexsym}
\usepackage{amsmath}
\usepackage{booktabs}
\usepackage{multirow}

\usepackage[T1]{fontenc}

\usepackage[utf8]{inputenc}

\usepackage{microtype}

\title{Evaluation of Social Biases in Recent Large Pre-Trained Models}

\author{Swapnil Sharma \And Nikita Anand \And Kranthi Kiran GV \And Alind Jain }

\begin{document}
\maketitle
\begin{abstract}
Large pre-trained language models are widely used in the community. These models are usually trained on unmoderated and unfiltered data from open sources like the internet. Due to this, biases that we see in platforms online which are a reflection of those in society are in turn captured and learnt by these models. These models are deployed in applications that affect millions of people and their inherent biases are harmful to the targeted social groups. In this work, we study the general trend in bias reduction as newer pre-trained models are released. Three recent models ( ELECTRA, DeBERTa and DistilBERT) are chosen and evaluated against two bias benchmarks, StereoSet and CrowS-Pairs. They are compared to the baseline of BERT using the associated metrics. We explore whether as advancements are made and newer, faster, lighter models are released: are they being developed responsibly such that their inherent social biases have been reduced compared to their older counterparts? The results are compiled and we find that all the models under study do exhibit biases but have generally improved as compared to BERT. Link to the code is available on Github.\footnote{ \href{https://github.com/KranthiGV/bias-evaluation}{https://github.com/KranthiGV/bias-evaluation}}
\end{abstract}

\section{Introduction}

Given the popularity of large pre-trained language models and their widespread use in the community, it is pertinent to evaluate their inherent biases. These biases come about because these models (for example, BERT \citep{devlin-etal-2019-bert} and BERT based models, ALBERT \citep{Lan2020ALBERT:}, RoBERTa \citep{DBLP:journals/corr/abs-1907-11692}) are trained on large sets of often freely available and unmoderated text data from sources such as the internet. 

Due to this, gender, race, religious and other social biases that we see in the real world are often translated into the models. Since these models are deployed in applications that are used by a large number of people, this bias is harmful. For example, these models are commonly used in content moderation tasks on social media platforms online. If the model is unfairly biased against certain social groups- minorities, marginalized people and those sections of society that are discriminated against, then they would be adversely affected by the application. So it is of utmost importance to ensure that models are trained in such a way that the bias is mitigated or to de-bias them.

There are multiple studies \citep{DBLP:conf/acl/MeadePR22,DBLP:journals/corr/abs-2109-05704,bhardwaj2021investigating,DBLP:journals/corr/abs-1906-07337} on social biases in BERT and older language models. We extend this and evaluate the relatively newer models: ELECTRA \citep{DBLP:journals/corr/abs-2003-10555}, DistilBERT \citep{DBLP:journals/corr/abs-1910-01108} and DeBERTa \citep{DBLP:journals/corr/abs-2006-03654}. These models have gained popularity due to factors such as better performance, low computer requirements and parameter efficiency. Each model is evaluated against two bias evaluation datasets, StereoSet \citep{nadeem-etal-2021-stereoset} and Crowdsourced Stereotype Pairs, CrowS-Pairs \citep{DBLP:journals/corr/abs-2010-00133}. StereoSet accounts for four different biases that are based on- gender, occupation, race and religion. In addition to these, CrowS-Pairs also has sexual orientation, age, nationality, disability and physical appearance. 

We aim to answer the question: as technology advances and newer pre-trained language models are released, are they being done responsibly such that their inherent social biases have been reduced? We believe that while making improvements to speed, size, efficiency and performance of models is at the forefront of research, equal importance needs to be given to mitigating their biases given their impact in the real world. 

The three newer models are run against these benchmarks and compared to the baseline of BERT. This gives us a general trend in whether bias is reduced as advancements in language models are being made. We find that all the models exhibit bias, but the three recent models perform better on the benchmarks than BERT. Models have varying results across different bias categories and we attempt to analyse their behaviour.

\section{Related work}

The empirical survey conducted in \citet{DBLP:conf/acl/MeadePR22} studied five different debiasing techniques Counterfactual Data Augmentation \citep{zmigrod-etal-2019-counterfactual}, Dropout \citep{DBLP:journals/corr/abs-2010-06032}, Iterative Nullspace Projection \citep{ravfogel-etal-2020-null}, Self-Debias \citep{10.1162/tacl_a_00434} and SentenceDebias \citep{liang-etal-2020-towards} on racial, religious and gender biases. They applied these techniques on BERT, RobERTa, ALBERT and GPT-2 \citep{radford2019language} and evaluated them against three bias benchmarks (StereoSet, CrowS-Pairs and Sentence Encoder Association Test \citep{may-etal-2019-measuring}. They report which technique is best at debiasing and whether these techniques hamper a model’s language modeling ability and performance on downstream Natural Language Understanding tasks.

Crowdsourced Stereotype Pairs, CrowS-Pairs \citep{DBLP:journals/corr/abs-2010-00133} is a crowdsourced bias evaluation dataset that deals with nine different types of social bias. Each sample consists of a pair of sentences- a stereotype and an anti-stereotype. This high quality dataset provides us a good way to gain insights on varying types of bias in recent models.

An earlier bias evaluation benchmark StereoSet \citep{nadeem-etal-2021-stereoset} deals with four types of biases but on a larger scale. They introduce the Context Association Test that "measures the language modeling ability as well as the stereotypical bias of pretrained language models". This work provides us with a large amount of data, though it does not cover as many biases as CrowS-Pairs. StereoSet has two kinds of tasks, intrasentence, which is sentence level and intersentence, which is discourse level. The intrasentence task has a fill-in-the-blank style structure describing a group where there is three options for the attribute, a stereotype, an anti-stereotype and an unrelated one. The intersentence task gives a context about the target group and the second contains the attribute. CrowS-Pairs and StereoSet's intrasentence task both have the similar structure of two sentences, the first being more biased that the second. The difference is that in the former the group being discussed changes between the two sentences while in the latter, the attribute assigned to the group changes. CrowS-Pairs conducted a validation study and came out rated much higher at 80\% compared to StereoSet's 62\%. Since CrowS-Pairs improves only on StereoSet's intrasentence task, we use it alongside the intersentence task.

\citet{bhardwaj2021investigating} does a deep-dive into gender bias in BERT. They examine the bias brought about by BERT in five downstream tasks based on sentiment and emotion intensity in Tweets. It is pointed out that intuitively and ideally the words that are gender indicative should matter little for the model's predictions but in actuality, they turn out to influence the predictions heavily. A layer-level solution is proposed wherein directions that encode gender are picked out and embeddings in the same are removed to largely reduce bias in the downstream tasks. This is one of the first studies that analyses bias in BERT in this manner.

A different perspective to the bias problem is offered by \citet{DBLP:journals/corr/abs-2109-05704} where the language dependent nature of bias is reported. Ethnic bias is studied across multiple languages using various monolingual versions of BERT. They alleviate bias using two methods: using a multilingual model and aligning words in two multilingual models.

\section{Methodology}

The methodology consists of the following stages: (1) Data loading and preprocessing, (2) Model loading and (3) Evaluation. We replicate the evaluation pipeline in CrowS-Pairs and StereoSet.

In data loading and preprocessing stage, we load a particular bias benchmark dataset and process it to ingest it further into the pipeline. This includes tokenization according the pre-trained model chosen. Evaluation consists of computing the metrics after running the given model on the bias benchmark dataset chosen. 

\subsection{Choice of Models}
Our choice of recent pre-trained models largely depends on the Hugging Face \citep{wolf-etal-2020-transformers} downloads and citation count as a guide to how popular and used they are. We compare them to a BERT baseline.

\paragraph{DistilBERT} retains 97\% of BERT's language understanding capabilities while being 40\% smaller and 60\% faster. This model has been downloaded from the Hugging Face library over 5 million times in the last month alone and has about 1700 citations.

\paragraph{DeBERTa} consistently outperforms a BERT based model RoBERTa on many NLP tasks. It has been downloaded over 1 million times over the last month which indicates its popularity in the community. 

\paragraph{ELECTRA} shows significant improvement in contextual representations over BERT given the same data and compute using a novel pre-training technique. ELECTRA (small, base and large) had over 600,000 downloads and 1270 citations.

\subsection{Datasets}

\subsubsection{StereoSet}

Stereoset dataset contains about 16,995 test sentences evaluating biases in four domains which are race, profession, gender and religion. Nearly 85\% of the examples are of profession or race bias categories and religion constitutes merely 4\% of the total.


The dataset contains two kinds of context association tests (CAT): Intersentence (sentence level) and Intrasentence (discourse level), both of which contain a Context, a Stereotype, an Anti-Stereotype and an Unrelated component. 

The evaluation metrics defined in \citet{nadeem-etal-2021-stereoset} are Language Modeling Score (lms), Stereotype Score (ss) and a combination of the two, Idealized CAT Score (icat). As per the paper,  lms of a target term is "the percentage of instances in which a language model prefers the
meaningful over meaningless association" and ss is "the percentage of examples in which a model prefers a stereotypical association over an anti-stereotypical association". The combination of the two gives form to icat which is mathematically described as follows,
\begin{align}
icat = \frac{lms \times min(ss, 100-ss)}{50} 
\end{align}
This metric captures the ability of a model to remain unbiased while performing language modeling extremely well.
\subsubsection{CrowS-Pairs}

CrowS-Pairs is a well curated, crowdsourced dataset that covers nine types of social biases. Each example comes in pairs of a stereotype and a corresponding anti-stereotype. To do this, crowdworkers were asked to write a stereotyping statement about a targeted group and then make minimal changes so that the target group is now an advantaged group instead. In this process, the words that do not change are called unmodified tokens and the ones that do are called modified tokens. They performed a series of checks to ensure the data is validated. The resulting dataset has 1508 examples, with the race category comprising nearly third of the total. 73.5\% of the data belongs to either the race, gender, occupation or nationality categories. They introduce their own metric, given by the \emph{pseudo-log-likelihood} scoring,
\begin{align}
score(S) = \sum_{i=0}^{|C|} \log P(u_i \in U | U_{\char`\\u_i}, M, \theta)
\end{align}

In this equation, U is the set of unmodified tokens, M is the set of modified tokens and S is the union of the two.

\begin{table*}[t!]
\small \centering
    \begin{tabular}{lccccc}
    \toprule
    \bf & \bf BERT$_\emph{base} (baseline)$ & \bf DistilBERT$_\emph{base}$ & \bf DeBERTa$_\emph{base}$ & \bf ELECTRA$_\emph{base}$\\
    \midrule
    CrowS-Pairs & 60.48 & 56.83 & 52.59 & 56.76 \\
    CrowS-Pairs-\emph{stereo} & 61.09 & 57.33 & 53.85 & 56.98 \\
    CrowS-Pairs-\emph{antistereo} & 56.88 & 54.13 & 45.87 & 55.76 \\
    \midrule
    \multicolumn{4}{l}{\emph{Bias Categories in CrowS-Pairs}} \\
    \midrule
    Race / Color & 58.14 & 53.29 & 56.01 & 57.95 \\
    Gender / Gender Identity & 58.02 & 54.58 & 43.51 & 52.67\\
    Socioeconomic status / Occupation & 59.88 & 55.81 & 55.23 & 61.63 \\
    Nationality & 62.89 & 54.09 & 55.97 & 53.46 \\
    Religion & 71.43 & 70.48 & 49.52 & 65.71 \\
    Age & 55.17 & 59.77 & 48.28 & 56.32 \\
    Sexual Orientation & 67.86 & 70.24 & 59.52 & 50.0 \\
    Physical Appearance & 63.49 & 55.56 & 53.97 & 53.97 \\
    Disability & 61.67 & 61.67 & 46.67 & 56.67 \\
    \bottomrule
    \end{tabular} 

\caption{\label{table:1}Model performances on CrowS-Pairs. Higher numbers indicate higher bias.}
\end{table*}

\begin{table*}[t!]
\small \centering
    \begin{tabular}{lll}
    \toprule
    \multirow{2}{*}{Low bias} & He could not believe his wife could be so callous.\\
    &He could not believe his husband could be so callous.\\
    \hline
    \multirow{2}{*}{High bias}&Gay people are really flamboyant and are always wearing ridiculous costumes.\\
    &Straight people are really flamboyant and are always wearing ridiculous costumes.\\
    \bottomrule
    \end{tabular} 

\caption{\label{table:3}Examples of sentence pairs in CrowS-Pairs on which ELECTRA exhibited low bias and high bias.}
\end{table*}

\section{Experiments}

In this section, we describe the process followed to evaluate the models and the experimental results.

\paragraph{Stereoset Set Up}
The task at hand in stereoset inter-sentences is to pick an attribute sentence give a target term sentence. This task is very similar to the Next Sentence Prediction (NSP) task that was one of the original BERT training tasks. To better analyze the model bias on varied tasks we analyze them on Stereoset inter-sentence dataset.

However, none of the models under consideration apart from BERT-baseline, contain a pretrained NSP classification head. So we decided to train the NSP heads on a subset of the Wikipedia corpus \citep{wikidump} for these models. Our NSP classification head achieves a 94.88\% on DistilBERT-\emph{base}, 96.18\% on BERT-\emph{base}, 95.18\% on DeBERTa-\emph{base} and 95.48\% on ELECTRA-\emph{base}.


\begin{table}[t!]
\small \centering
    \begin{tabular}{lcccc}
    \toprule
    \bf Bias & \bf Count & \bf lms & \bf ss & \bf icat \\
    \midrule
    \multicolumn{5}{c}{BERT$_\emph{base} (baseline)$} \\
    \midrule
    Gender & 726 & 89.26 & 59.09 & 73.03 \\
    Profession & 2481 & 84.70 & 63.48 & 61.86 \\
    Race & 2928 & 87.76 & 60.35 & 69.59 \\
    Religion & 234 & 91.03 & 61.54 & 70.02 \\
    Overall & 2123 & 86.86 & 61.47 & 66.99 \\
    \midrule
    \multicolumn{5}{c}{DistilBERT$_\emph{base}$}\\
    \midrule
    Gender & 726 & 87.56 & 56.78 & 75.70 \\
    Profession & 2481 & 86.50 & 58.25 & 72.24 \\
    Race & 2928 & 82.77 & 56.89 & 71.36 \\
    Religion & 234 & 86.29 & 62.02 & 65.55 \\
    Overall & 2123 & 84.92 & 57.59 & 72.04 \\
    \midrule
    \multicolumn{5}{c}{DeBERTa$_\emph{base}$} \\
    \midrule
    Gender & 726 & 77.49 & 8.46 & 13.11 \\
    Profession & 2481 & 80.25 & 24.03 & 38.57 \\
    Race & 2928 & 81.15 & 23.57 & 38.26 \\
    Religion & 234 & 82.49 & 18.69 & 30.83 \\
    Overall & 2123 & 80.40 & 21.65 & 34.81 \\
    \midrule
    \multicolumn{5}{c}{ELECTRA$_\emph{base}$}\\
    \midrule
    Gender & 726 & 90.79 & 56.60 & 78.81 \\
    Profession & 2481 & 87.98 & 60.60 & 69.33 \\
    Race & 2928 & 88.37 & 57.14 & 75.75 \\
    Religion & 234 & 97.52 & 53.84 & 90.02 \\
    Overall & 2123 & 88.87 & 58.26 & 74.19 \\
    \bottomrule
    \end{tabular} 

\caption{\label{table:2}Model performances on StereoSet intersentences. Numbers greater than 50 indicate higher bias.}
\end{table}



\subsection{Results}

In Table \ref{table:1}, we observe that on the CrowS-Pairs benchmark dataset, all the models exhibit bias. BERT has the highest bias score while models such as DeBERTa and ELECTRA perform better. DeBERTa has a metric score of $52.59$ which is the lowest score among the models we studied in this work. We also observe that the models do not demonstrate similar scores in each of the bias categories of CrowS-Pairs dataset. For instance, DistilBERT exhibits higher bias on Religion ($70.48$) and Sexual Orientation ($70.24$) than other categories such as Gender ($54.58$), Race ($53.29$) and others. ELECTRA, however, exhibits lower bias on Religion ($65.71$) than DistilBERT.

Table \ref{table:3} presents examples of sentences on which ELECTRA model exhibited low and high bias. In the example on which it exhibited higher bias: "Gay people are flamboyant\ldots", it is likely due to the presence of words such as "flamboyant" which may have been typically found along with the word "Gay" on the training corpus. The examples on which it exhibited low bias, it could be likely due to the words being quite general "wife" and "husband".

In Table \ref{table:2}, we observe the outcome running BERT against StereoSet. Similar to CrowS-Pair evaluation, we observe that models have different scores in each bias categories. The Language Modeling Score was slightly higher in the Religion category as compared to the others. 

DeBERTa in both CrowS-Pairs and stereoset shows inclination to anti-stereotype. The anti-stereotype result seems more pronounced in case of  intersentences. We hypothesize that this effect is more vividly visible because we have fine tuned DeBERTa for NSP task on Wikipedia Corpus. Since DeBERTa has far less trainable parameters as compared to other models, it ends up learning anti-stereotype behaviour from strictly monitored Wikipedia content.



\section{Ethical Considerations}

The study we conduct in this work is meant to indicate the progress made in mitigating social biases in recent models. A model performing well on these bias benchmarks alone should not be interpreted as it being unbiased. 

StereoSet only considers a binary definition of gender. Going forward we believe it is crucial for studies on gender bias to be non-binary.

The benchmarks employed only North American crowdworkers so the stereotypes are narrow in that regard and do not represent biases across cultures. 

\section{Collaboration Statement}
Kranthi implemented the CrowS-Pairs evaluation and Swapnil worked on StereoSet. Nikita worked on Introduction, Literature Review, Methodology and experiments with SEAT (not included in this report). We worked together to analyse the results. Alind worked on Datasets and everyone worked on the presentation.


\bibliography{anthology,custom}

\begin{thebibliography}{21}
\expandafter\ifx\csname natexlab\endcsname\relax\def\natexlab#1{#1}\fi

\bibitem[{Ahn and Oh(2021)}]{DBLP:journals/corr/abs-2109-05704}
Jaimeen Ahn and Alice Oh. 2021.
\newblock Mitigating language-dependent ethnic bias in bert.
\newblock In \emph{Proceedings of the 2021 Conference on Empirical Methods in
  Natural Language Processing}, pages 533--549.

\bibitem[{Beutel et~al.(2020)Beutel, Chi, Pavlick, Pitler, Tenney, Chen,
  Webster, Petrov, and Wang}]{DBLP:journals/corr/abs-2010-06032}
Alex Beutel, Ed~H. Chi, Ellie Pavlick, Emily~Blythe Pitler, Ian Tenney, Jilin
  Chen, Kellie Webster, Slav Petrov, and Xuezhi Wang. 2020.
\newblock \href {https://arxiv.org/abs/2010.06032} {Measuring and reducing
  gendered correlations in pre-trained models}.
\newblock Technical report.

\bibitem[{Bhardwaj et~al.(2021)Bhardwaj, Majumder, and
  Poria}]{bhardwaj2021investigating}
Rishabh Bhardwaj, Navonil Majumder, and Soujanya Poria. 2021.
\newblock Investigating gender bias in bert.
\newblock \emph{Cognitive Computation}, 13(4):1008--1018.

\bibitem[{Clark et~al.(2020)Clark, Luong, Le, and
  Manning}]{DBLP:journals/corr/abs-2003-10555}
Kevin Clark, Minh-Thang Luong, Quoc~V. Le, and Christopher~D. Manning. 2020.
\newblock \href {https://openreview.net/pdf?id=r1xMH1BtvB} {{ELECTRA}:
  Pre-training text encoders as discriminators rather than generators}.
\newblock In \emph{ICLR}.

\bibitem[{Devlin et~al.(2019)Devlin, Chang, Lee, and
  Toutanova}]{devlin-etal-2019-bert}
Jacob Devlin, Ming-Wei Chang, Kenton Lee, and Kristina Toutanova. 2019.
\newblock \href {https://doi.org/10.18653/v1/N19-1423} {{BERT}: Pre-training of
  deep bidirectional transformers for language understanding}.
\newblock In \emph{Proceedings of the 2019 Conference of the North {A}merican
  Chapter of the Association for Computational Linguistics: Human Language
  Technologies, Volume 1 (Long and Short Papers)}, pages 4171--4186,
  Minneapolis, Minnesota. Association for Computational Linguistics.

\bibitem[{Foundation()}]{wikidump}
Wikimedia Foundation.
\newblock \href {https://dumps.wikimedia.org} {Wikimedia downloads}.

\bibitem[{He et~al.(2020)He, Liu, Gao, and
  Chen}]{DBLP:journals/corr/abs-2006-03654}
Pengcheng He, Xiaodong Liu, Jianfeng Gao, and Weizhu Chen. 2020.
\newblock Deberta: Decoding-enhanced bert with disentangled attention.
\newblock In \emph{International Conference on Learning Representations}.

\bibitem[{Kurita et~al.(2019)Kurita, Vyas, Pareek, Black, and
  Tsvetkov}]{DBLP:journals/corr/abs-1906-07337}
Keita Kurita, Nidhi Vyas, Ayush Pareek, Alan~W Black, and Yulia Tsvetkov. 2019.
\newblock Measuring bias in contextualized word representations.
\newblock In \emph{Proceedings of the First Workshop on Gender Bias in Natural
  Language Processing}, pages 166--172.

\bibitem[{Lan et~al.(2020)Lan, Chen, Goodman, Gimpel, Sharma, and
  Soricut}]{Lan2020ALBERT:}
Zhenzhong Lan, Mingda Chen, Sebastian Goodman, Kevin Gimpel, Piyush Sharma, and
  Radu Soricut. 2020.
\newblock \href {https://openreview.net/forum?id=H1eA7AEtvS} {Albert: A lite
  bert for self-supervised learning of language representations}.
\newblock In \emph{International Conference on Learning Representations}.

\bibitem[{Liang et~al.(2020)Liang, Li, Zheng, Lim, Salakhutdinov, and
  Morency}]{liang-etal-2020-towards}
Paul~Pu Liang, Irene~Mengze Li, Emily Zheng, Yao~Chong Lim, Ruslan
  Salakhutdinov, and Louis-Philippe Morency. 2020.
\newblock \href {https://doi.org/10.18653/v1/2020.acl-main.488} {Towards
  debiasing sentence representations}.
\newblock In \emph{Proceedings of the 58th Annual Meeting of the Association
  for Computational Linguistics}, pages 5502--5515, Online. Association for
  Computational Linguistics.

\bibitem[{Liu et~al.(2019)Liu, Ott, Goyal, Du, Joshi, Chen, Levy, Lewis,
  Zettlemoyer, and Stoyanov}]{DBLP:journals/corr/abs-1907-11692}
Yinhan Liu, Myle Ott, Naman Goyal, Jingfei Du, Mandar Joshi, Danqi Chen, Omer
  Levy, Mike Lewis, Luke Zettlemoyer, and Veselin Stoyanov. 2019.
\newblock \href {http://arxiv.org/abs/1907.11692} {Roberta: {A} robustly
  optimized {BERT} pretraining approach}.
\newblock \emph{CoRR}, abs/1907.11692.

\bibitem[{May et~al.(2019)May, Wang, Bordia, Bowman, and
  Rudinger}]{may-etal-2019-measuring}
Chandler May, Alex Wang, Shikha Bordia, Samuel~R. Bowman, and Rachel Rudinger.
  2019.
\newblock \href {https://doi.org/10.18653/v1/N19-1063} {On measuring social
  biases in sentence encoders}.
\newblock In \emph{Proceedings of the 2019 Conference of the North {A}merican
  Chapter of the Association for Computational Linguistics: Human Language
  Technologies, Volume 1 (Long and Short Papers)}, pages 622--628, Minneapolis,
  Minnesota. Association for Computational Linguistics.

\bibitem[{Meade et~al.(2022)Meade, Poole{-}Dayan, and
  Reddy}]{DBLP:conf/acl/MeadePR22}
Nicholas Meade, Elinor Poole{-}Dayan, and Siva Reddy. 2022.
\newblock \href {https://aclanthology.org/2022.acl-long.132} {An empirical
  survey of the effectiveness of debiasing techniques for pre-trained language
  models}.
\newblock In \emph{Proceedings of the 60th Annual Meeting of the Association
  for Computational Linguistics (Volume 1: Long Papers), {ACL} 2022, Dublin,
  Ireland, May 22-27, 2022}, pages 1878--1898. Association for Computational
  Linguistics.

\bibitem[{Nadeem et~al.(2021)Nadeem, Bethke, and
  Reddy}]{nadeem-etal-2021-stereoset}
Moin Nadeem, Anna Bethke, and Siva Reddy. 2021.
\newblock \href {https://doi.org/10.18653/v1/2021.acl-long.416} {{S}tereo{S}et:
  Measuring stereotypical bias in pretrained language models}.
\newblock In \emph{Proceedings of the 59th Annual Meeting of the Association
  for Computational Linguistics and the 11th International Joint Conference on
  Natural Language Processing (Volume 1: Long Papers)}, pages 5356--5371,
  Online. Association for Computational Linguistics.

\bibitem[{Nangia et~al.(2020)Nangia, Vania, Bhalerao, and
  Bowman}]{DBLP:journals/corr/abs-2010-00133}
Nikita Nangia, Clara Vania, Rasika Bhalerao, and Samuel Bowman. 2020.
\newblock Crows-pairs: A challenge dataset for measuring social biases in
  masked language models.
\newblock In \emph{Proceedings of the 2020 Conference on Empirical Methods in
  Natural Language Processing (EMNLP)}, pages 1953--1967.

\bibitem[{Radford et~al.(2019)Radford, Wu, Child, Luan, Amodei, and
  Sutskever}]{radford2019language}
Alec Radford, Jeff Wu, Rewon Child, David Luan, Dario Amodei, and Ilya
  Sutskever. 2019.
\newblock Language models are unsupervised multitask learners.

\bibitem[{Ravfogel et~al.(2020)Ravfogel, Elazar, Gonen, Twiton, and
  Goldberg}]{ravfogel-etal-2020-null}
Shauli Ravfogel, Yanai Elazar, Hila Gonen, Michael Twiton, and Yoav Goldberg.
  2020.
\newblock \href {https://doi.org/10.18653/v1/2020.acl-main.647} {Null it out:
  Guarding protected attributes by iterative nullspace projection}.
\newblock In \emph{Proceedings of the 58th Annual Meeting of the Association
  for Computational Linguistics}, pages 7237--7256, Online. Association for
  Computational Linguistics.

\bibitem[{Sanh et~al.(2019)Sanh, Debut, Chaumond, and
  Wolf}]{DBLP:journals/corr/abs-1910-01108}
Victor Sanh, Lysandre Debut, Julien Chaumond, and Thomas Wolf. 2019.
\newblock Distilbert, a distilled version of {BERT:} smaller, faster, cheaper
  and lighter.
\newblock In \emph{Fifth Workshop on Energy Efficient Machine Learning and
  Cognitive Computing - NeurIPS Edition}.

\bibitem[{Schick et~al.(2021)Schick, Udupa, and
  Schütze}]{10.1162/tacl_a_00434}
Timo Schick, Sahana Udupa, and Hinrich Schütze. 2021.
\newblock \href {https://doi.org/10.1162/tacl_a_00434} {{Self-Diagnosis and
  Self-Debiasing: A Proposal for Reducing Corpus-Based Bias in NLP}}.
\newblock \emph{Transactions of the Association for Computational Linguistics},
  9:1408--1424.

\bibitem[{Wolf et~al.(2020)Wolf, Debut, Sanh, Chaumond, Delangue, Moi, Cistac,
  Rault, Louf, Funtowicz, Davison, Shleifer, von Platen, Ma, Jernite, Plu, Xu,
  Le~Scao, Gugger, Drame, Lhoest, and Rush}]{wolf-etal-2020-transformers}
Thomas Wolf, Lysandre Debut, Victor Sanh, Julien Chaumond, Clement Delangue,
  Anthony Moi, Pierric Cistac, Tim Rault, Remi Louf, Morgan Funtowicz, Joe
  Davison, Sam Shleifer, Patrick von Platen, Clara Ma, Yacine Jernite, Julien
  Plu, Canwen Xu, Teven Le~Scao, Sylvain Gugger, Mariama Drame, Quentin Lhoest,
  and Alexander Rush. 2020.
\newblock \href {https://doi.org/10.18653/v1/2020.emnlp-demos.6} {Transformers:
  State-of-the-art natural language processing}.
\newblock In \emph{Proceedings of the 2020 Conference on Empirical Methods in
  Natural Language Processing: System Demonstrations}, pages 38--45, Online.
  Association for Computational Linguistics.

\bibitem[{Zmigrod et~al.(2019)Zmigrod, Mielke, Wallach, and
  Cotterell}]{zmigrod-etal-2019-counterfactual}
Ran Zmigrod, Sabrina~J. Mielke, Hanna Wallach, and Ryan Cotterell. 2019.
\newblock \href {https://doi.org/10.18653/v1/P19-1161} {Counterfactual data
  augmentation for mitigating gender stereotypes in languages with rich
  morphology}.
\newblock In \emph{Proceedings of the 57th Annual Meeting of the Association
  for Computational Linguistics}, pages 1651--1661, Florence, Italy.
  Association for Computational Linguistics.

\end{thebibliography}
\bibliographystyle{acl_natbib}

\clearpage
\appendix

\section{Appendix}
\label{sec:appendix}

In Table \ref{tab:acc} we show the models evaluated on StereoSet Intrasentence dataset. CrowS-Pairs found errors in the StereoSet Intrasentences and so we ask readers to be cautious of these results.  


\begin{table}[t!]
\small \centering
    \begin{tabular}{lcccc}
    \toprule
    \bf Bias & \bf Count & \bf lms & \bf ss & \bf icat \\
    \midrule
    \multicolumn{5}{c}{BERT$_\emph{base} (Baseline)$} \\
    \midrule
    Gender & 765 & 85.97 & 63.56 & 62.65 \\
    Profession & 2430 & 82.72 & 61.16 & 64.26 \\
    Race & 2886 & 85.45 & 57.24 & 73.07 \\
    Religion & 237 & 88.46 & 55.13 & 79.39 \\
    Overall & 2106 & 84.59 & 59.45 & 68.60 \\
    \midrule
    \multicolumn{5}{c}{DistilBERT$_\emph{base}$}\\
    \midrule
    Gender & 765 & 86.43 & 59.76 & 69.56 \\
    Profession & 2430 & 83.50 & 64.06 & 60.02 \\
    Race & 2886 & 86.71 & 58.05 & 72.74 \\
    Religion & 237 & 89.13 & 61.98 & 67.78 \\
    Overall & 2106 & 85.55 & 60.70 & 67.24 \\
    \midrule
    \multicolumn{5}{c}{DeBERTa$_\emph{base}$} \\
    \midrule
    Gender & 765 & 54.41 & 50.54 & 53.83 \\
    Profession & 2430 & 44.06 & 49.64 & 43.74 \\
    Race & 2886 & 49.02 & 50.64 & 48.40 \\
    Religion & 237 & 51.17 & 57.06 & 43.95 \\
    Overall & 2106 & 47.90 & 50.49 & 47.43 \\
    \midrule
    \multicolumn{5}{c}{ELECTRA$_\emph{base}$}\\
    \midrule
    Gender & 765 & 57.35 & 48.19 & 55.28 \\
    Profession & 2430 & 58.47 & 46.19 & 54.00 \\
    Race & 2886 & 59.09 & 43.83 & 51.80 \\
    Religion & 237 & 60.57 & 45.98 & 55.70 \\
    Overall & 2106 & 58.69 & 45.36 & 53.24 \\
    \bottomrule
    \end{tabular} 

\caption{\label{tab:acc}StereoSet intrasentences results.}
\end{table}

\end{document}